\documentclass[conference]{IEEEtran}
\IEEEoverridecommandlockouts
\usepackage{cite}
\usepackage{amsmath,amssymb,amsfonts}
\usepackage{algorithmic}
\usepackage{bbold}  
\usepackage{comment}
\usepackage{graphicx}
\usepackage{textcomp}
\usepackage{xcolor}
\usepackage{hyperref}
\usepackage{siunitx}
\usepackage{tabularx}
\usepackage{booktabs}
\newcolumntype{C}{>{\centering\arraybackslash}X} % defined centered version of "X" column type:
\usepackage[colorinlistoftodos,prependcaption,textsize=tiny]{todonotes}
\def\BibTeX{{\rm B\kern-.05em{\sc i\kern-.025em b}\kern-.08em
    T\kern-.1667em\lower.7ex\hbox{E}\kern-.125emX}}

\usepackage[T1]{fontenc}
\usepackage{etoolbox}

\newcommand\copyrighttext{%
  \footnotesize \textcopyright 2023 IEEE. Personal use of this material is permitted.
  Permission from IEEE must be obtained for all other uses, in any current or future
  media, including reprinting/republishing this material for advertising or promotional
  purposes, creating new collective works, for resale or redistribution to servers or
  lists, or reuse of any copyrighted component of this work in other works.}

\newcommand\copyrightnotice{%
\begin{tikzpicture}[remember picture,overlay]
\node[anchor=south,yshift=10pt] at (current page.south) {\fbox{\parbox{\dimexpr\textwidth-\fboxsep-\fboxrule\relax}{\copyrighttext}}};
\end{tikzpicture}%
}
\makeatletter
\patchcmd{\maketitle}{\@fnsymbol}{\@alph}{}{}  % Footnote numbers from symbols to small letters
\makeatother

\title{The IMPTC Dataset: An Infrastructural Multi-Person Trajectory and Context Dataset}

\author{
    Manuel Hetzel\thanks{\textsuperscript{*} The authors contributed equally.}\textsuperscript{*}\thanks{\textsuperscript{1}The authors are with the Faculty of Engineering,
		University of Applied Sciences Aschaffenburg, Aschaffenburg, Germany. E-mail: \{manuel.hetzel, hannes.reichert, konrad.doll\}@th-ab.de.}\textsuperscript{1}
    \and
    Hannes Reichert\textsuperscript{*}\textsuperscript{1}
    \and
    Günther Reitberger\textsuperscript{*}\thanks{\textsuperscript{2}The authors are with FORWISS, University of Passau, Passau, Germany. E-mail: \{reitberg, fuchse\}@forwiss.uni-passau.de.}\textsuperscript{2}
    \and
    Erich Fuchs\textsuperscript{2}
    \and
    Konrad Doll\textsuperscript{1}
    \and
    Bernhard Sick\thanks{\textsuperscript{3}The author is with the Intelligent Embedded Systems Lab, University of Kassel, Kassel, Germany. E-mail: \{bsick\}@uni-kassel.de.}\textsuperscript{3}

    \thanks{This work results from the project AI Data Tooling supported by the German Federal Ministry for Economic Affairs and Climate Actions (BMWK), grant numbers, 19A20001L, 19A20001P, and 19A20001O.}

}

\begin{document}

\maketitle

%---- Copyright Notice
\copyrightnotice
%----

\begin{figure*}[h]
    \centering
    \includegraphics[clip,width=\textwidth]{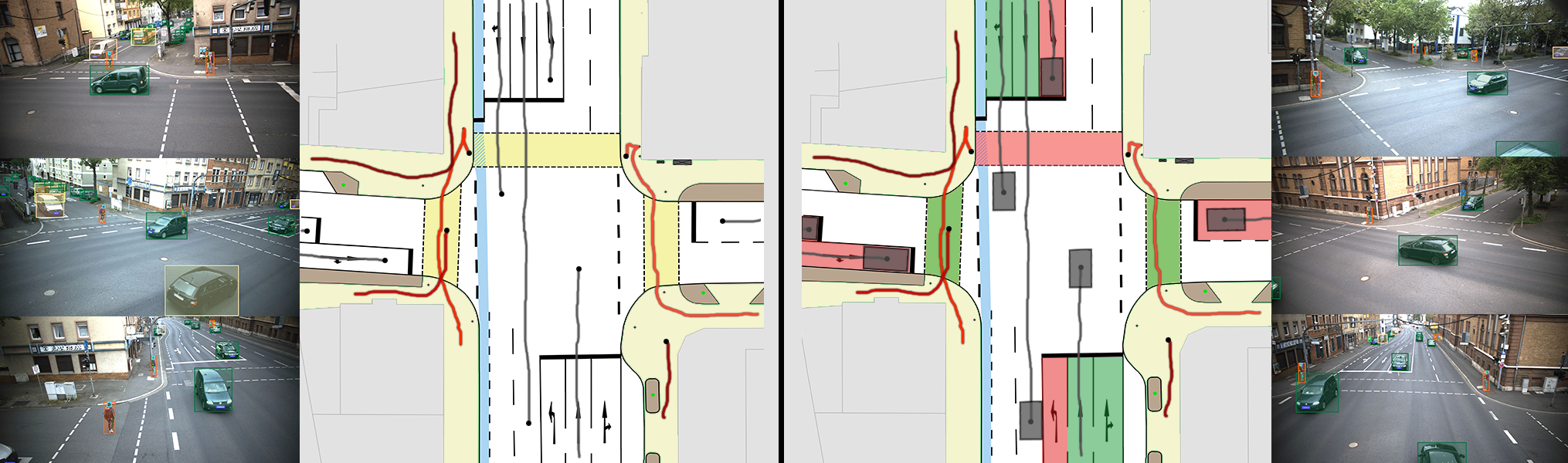}
    \caption{An exemplary scene presentation of road user trajectories within the IMPTC dataset. The figure illustrates all six camera perspectives with object detection results as colored bounding boxes and two top-view maps with road user trajectories. The left map only visualizes trajectory data, whereas the right map includes additional context information like traffic light signal status and 3D vehicle sizes. VRU tracks use red colors, and vehicles use grayish tones.}
    \label{fig:scene_preview}
\end{figure*}

\begin{abstract} 
Inner-city intersections are among the most critical traffic areas for injury and fatal accidents. Automated vehicles struggle with the complex and hectic everyday life within those areas. Sensor-equipped smart infrastructures, which can cooperate with vehicles, can benefit automated traffic by extending the perception capabilities of drivers and vehicle perception systems. Additionally, they offer the opportunity to gather reproducible and precise data of a holistic scene understanding, including context information as a basis for training algorithms for various applications in automated traffic. Therefore, we introduce the Infrastructural Multi-Person Trajectory and Context Dataset (IMPTC). We use an intelligent public inner-city intersection in Germany with visual sensor technology. A multi-view camera and LiDAR system perceives traffic situations and road users' behavior. Additional sensors monitor contextual information like weather, lighting, and traffic light signal status. The data acquisition system focuses on Vulnerable Road Users (VRUs) and multi-agent interaction. The resulting dataset consists of eight hours of measurement data. It contains over 2,500 VRU trajectories, including pedestrians, cyclists, e-scooter riders, strollers, and wheelchair users, and over 20,000 vehicle trajectories at different day times, weather conditions, and seasons. In addition, to enable the entire stack of research capabilities, the dataset includes all data, starting from the sensor-, calibration- and detection data until trajectory and context data. The dataset is continuously expanded and is available online for non-commercial research at \url{https://github.com/kav-institute/imptc-dataset}.
\end{abstract}

\begin{IEEEkeywords}
Dataset, Trajectories, Vulnerable Road Users, Intelligent Infrastructure, Machine Learning
\end{IEEEkeywords}

\section{\large Introduction}\label{sec_introduction}
Automated driving (AD) is a major goal and point of interest for the industry and the research community. In the recent past, data-driven approaches have been a thriving factor in making this goal reachable. Datasets naturally play a central role as the performance of the algorithms strongly depends on them. Famous examples are JAAD~\cite{jaad}, SHIFT~\cite{shift_2022}, and inD~\cite{inD}. It is not just about the datasets themselves but also about how to create datasets, how to be able to evaluate results, and especially how to make sure that the data developed for algorithms represent the real world. In the case of synthetic datasets, e.g., SHIFT or sets created with CARLA~\cite{carla_2017}, it must be ensured that synthetic data can overcome domain gaps to improve real-world applications. With our work, we want to contribute a piece to the picture of holistic algorithms towards AD. We provide a sensor setup at a public inner-city intersection with everyday traffic. The setup can cover the whole intersection and parts of the incoming streets. Our main focus lies in predicting future residence probabilities and investigating the behavior of VRUs to include them in the AD environment, especially in complex inner-city scenarios. 

For VRU location and behavioral prediction, trajectory and additional context data are essential. If only trajectories are used, then the prediction relies on the past observed motion of the VRU to predict future locations. These algorithms react to an action already in progress instead of anticipating it. Additional information, e.g., VRU body poses and hand gestures or traffic light signal status, can help to improve the reliability and precision of VRU predictions. Many datasets do not provide this; for example, CityScapes~\cite{cityscapes_2016} or NuScenes~\cite{nuscenes} provides too short sequences of VRUs, making it more difficult to determine their intentions profoundly and especially to include group or context-induced behavior. Other datasets like JAAD or Euro-PVI (PVI)~\cite{euro_pvi} provide additional context information to improve scene understanding. We introduce an extensive real-world dataset, Infrastructural Multi-Person Trajectory and Context Dataset (IMPTC), providing complete scene understanding with additional context information and top-to-bottom data availability for VRU intention prediction modeling created by our infrastructural setup. With the publication of the dataset, we want to support research on safety validation for AD, VRU intention prediction, and other topics which rely on naturalistic VRU trajectories.

We use context as a superordinate term to describe all types of information regarding VRU behavior influencing road safety. That includes VRU attributes like gender, age, body pose, hand gesture, and view direction, just as static and dynamic conditions like traffic light signals, traffic rules, maps, weather, lightning, and interactions with other road users. 

In~\autoref{sec_state_of_the_art}, we give an overview of existing intelligent intersections and a summary of available VRU trajectory datasets. Afterward,~\autoref{sec_system_setup} presents our public research intersection, including sensor setup, data acquisition, and post-processing methods. Next, in~\autoref{sec_dataset}, we will detail the IMPTC dataset and compare it with currently available datasets. Finally,~\autoref{sec_conclusion} will summarize our work.
\section{\large Related Work}
\label{sec_state_of_the_art}

This chapter references prior work in two related topic areas: intelligent infrastructure systems, including datasets for VRU safety (\autoref{sec_intelligent_intersections}) and other existing methods for recording publicly available multi-VRU trajectory datasets (\autoref{sec_vru_trajectory_datasets}).

%-----------------------------------------------------------------------------------------------
\subsection{Intelligent Intersections}
\label{sec_intelligent_intersections}

There are multiple intelligent intersections used for road safety research applications. However, most research focuses on high-level traffic flow understanding and optimization~\cite{survey}. In contrast, some research targets VRU safety topics requiring high-resolution sensing technologies. 

\textbf{In Aschaffenburg}, Germany, a research intersection was introduced in 2012 for the German $\textit{Ko-PER}$~\cite{koper} and $\textit{DeCoInt}^2$~\cite{decoint} projects. A precise 90-degree stereo camera setup using two gray-scale full HD cameras has been used to detect and track VRU behavior focusing on one corner of the intersection. $\textit{Ko-PER}$ investigates prediction behavior models for VRUs crossing the street, $\textit{DeCoInt}^2$ covers VRU intention detection under the cooperative aspect between intelligent infrastructure and mobile research vehicles. For motion anticipation, \hyphenation{Reitberger}Reitberger et al.~\cite{reitberger} provided a cooperative tracking algorithm for cyclists, and Bieshaar et al.~\cite{bieshaar} used Convolutional Neural Networks to detect starting movements of cyclists. Zernetsch et al.~\cite{zernensch} developed a probabilistic VRU trajectory forecasting method. Kress et al.~\cite{kress} used this sensor setup as a reference to evaluate a human keypoint detection model deployed to a mobile research vehicle. It is worth mentioning that this sensor setup and the knowledge from the $\textit{Ko-PER}$ and $\textit{DeCoInt}^2$ projects were utilized to develop the novel proposed sensor setup. Regarding publicly available data, the $\textit{Ko-PER}$ project only provides a small amount of trajectory data for public research~\cite{koper_data}. The dataset contains VRUs and vehicles with 340 trajectories extracted from less than one hour of recordings. In contrast, $\textit{DeCoInt}^2$ released a pedestrian and cyclist trajectory dataset~\cite{decoint_dataset} extracted from the same intersection containing 2,700 VRU trajectories. Furthermore, a cyclist trajectory dataset is provided by Bieshaar et al.~\cite{bieshaar} with 84 sequences recorded by cellphone GPS data.

\textbf{In Braunschweig}, Germany, a comparable research intersection serves as a field instrument for detecting and assessing traffic behavior~\cite{dlr}. The intersection can provide trajectory data of road users acquired by multi-modal sensor setups. Mono cameras and radars are utilized for the 3D detection of vehicles. For VRU detection, multiple binocular stereo camera setups facing the pedestrian crossings are used. From this intersection, no data is publicly available.

\textbf{In Auburn Hills}, Michigan, Continental operates two intelligent intersections in public use~\cite{continental}. The systems improve traffic flow, reduce pollution, and increase the intersection’s safety by communicating hidden dangers to approaching connected vehicles and pedestrians. Camera and radar sensors create an environment model providing information about road users, traffic infrastructure, and static objects to connected vehicles using infrastructure-to-everything (I2X) communication. Continental does not provide publicly available data from its intelligent intersections.

\textbf{In Ulm}, Germany, another research intersection is located~\cite{ulm}. A combination of camera, LiDAR, and radar sensors creates road user trajectory data. In addition to the core intersection area, the sensors cover all three approaching streets for several hundred meters. Therefore, the general object tracking area is more extensive than all previously described research intersections. Just like in Braunschweig, no data is publicly available for researchers.

Besides intelligent infrastructures, other recording areas and techniques are used to create VRU trajectory datasets. The following section will present recent works.

%-----------------------------------------------------------------------------------------------
\subsection{VRU Trajectory Datasets}
\label{sec_vru_trajectory_datasets}

Several VRU trajectory datasets have been introduced and published in recent years. Different recording techniques, contents, scopes, and target research applications differentiate these sets. In terms of recording techniques, one can group them into three sub-classes, drone-based, vehicle-based, and stationary-based. The main focus of many datasets is trajectory prediction, but within the last couple of years, more datasets focusing on behavior and intention prediction topics have been published. The following will give an overview of existing and publicly available datasets. 

Drones have become very popular for road user trajectory recording within the last few years. A drone equipped with a camera monitors a specific environment on the ground. In most cases, the drone holds a static position up to 100 meters above the target area for recording. The inD~\cite{inD}, rounD~\cite{rounD}, HighD~\cite{highD}, SSD~\cite{ssd}, and CITR+DUT~\cite{citr} datasets use drones to acquire trajectories from critical intersections, public places of interest, or highways. This method is very flexible and enables fast data recording capabilities. However, typical roadside occlusions do not exist because of the top view. In return, useful context information is lost, such as traffic light signals, VRU body poses and gestures, vehicle flashing lights, and object heights. Furthermore, the trajectory precision of tiny objects like VRUs is error-prone, and drones can only operate in good weather conditions. Therefore, all drone datasets are recorded in calm and sunny weather. Rain, snowfall, or wind can affect the behavior of road users resulting in different behavior, which these sets do not cover. All previously mentioned drone-based datasets are publicly available.

Sensor-equipped mobile research vehicles are another popular way to record environment perception data. Cityscapes~\cite{cityscapes_2016} is among the first high-quality and extensive vehicle-based datasets. Followed by NuScenes~\cite{nuscenes}, Waymo Open Dataset~\cite{waymo_prediction}, BDD100K~\cite{bdd100k}, and ONCE~\cite{once}. In 2022 SHIFT~\cite{shift_2022} was introduced as a full synthetic dataset with over 2 million annotated frames. These vehicle-based datasets cover all types of road typologies and users. Therefore VRU related scenes must be filtered explicitly. Due to the amount of recorded data, only some frames are annotated, reducing the sample rate. Cityscapes uses a 17 Hz camera recording frame rate; every 20th frame is annotated, resulting in a less than 1 Hz sample rate. NuScenes has a frame rate of 2 Hz. Drone-based and infrastructure-based datasets provide much higher frame rates, resulting in a more precise mapping of the ongoing situation and VRU behavior. Furthermore, the sensors' point of view is vulnerable to occlusions. In contrast, the Pedestrian Intention Estimation Dataset (PIE)~\cite{pie} and the PVI datasets focuses on VRU behavior. All scenes include VRUs with additional labels to describe their behavior and the possible intention to cross the street. In-vehicle cameras depend on good weather and lighting conditions. Therefore PIE only includes scenes in sunny and calm weather. Moreover, only VRU and ego-vehicle trajectories are included, and no other road users are considered. All previously mentioned vehicle-based datasets are publicly available.

In contrast to intelligent intersections, there are other stationary mounting areas to capture trajectory data. The datasets from ETH~\cite{eth}, UCY~\cite{ucy}, ATC~\cite{atc}, and GC~\cite{gc} use stationary mounted sensors on top of or inside buildings. These datasets cover crowded areas with high VRU activities targeting human-to-human interactions. However, the datasets are not recorded under real-world traffic scenarios and do not focus on road traffic situations. For example, the ATC dataset consists of pedestrian trajectories in a shopping center, and in GC the main hall of the grand central station in New York is monitored. All previously presented stationary and drone- and vehicle-based datasets provide trajectory data of VRUs. However, the datasets vary in scope and research targets. Beyond that, only some datasets provide additional context data. In the case of VRU intention detection, trajectories are one of many parameters that need to be considered. Additional context information is necessary to create reliable VRU behavior predictions. Recent research demonstrates that additional context can significantly improve VRU trajectory forecasting. Rasouli et al.~\cite{pie} used local context, e.g., curbs or crosswalks, and VRU appearance, e.g., gestures, to create a pedestrian intention estimation model. The model's outcome is used with trajectory data as additional information for VRU trajectory prediction. Kress et al.~\cite{kress} demonstrated that human body-pose information helps to improve VRU intention detection. Instead of using one real-world coordinate representing a VRU's 3D location, 17 body joints are used. That enables a more precise presentation of VRU's posture and gestures. The lack of context and the imbalance of presented datasets is a gap for ongoing research in VRU intention prediction. We introduce our research intersection and the IMPTC dataset to address this issue.

\section{\large System Setup}
\label{sec_system_setup}

In this chapter, we introduce our research intersection. In~\autoref{sec_research_intersection}, we present the topology of the intersection, followed by~\autoref{sec_detection_classification} detailing the object detection and classification process. Next,~\autoref{sec_tracking_postprocessing} explains the object assignment and tracking, followed by~\autoref{sec_context_data} describing included context data. Finally,~\autoref{sec_format_tools} describes the dataset format.

%-----------------------------------------------------------------------------------------------
\subsection{Research Intersection}
\label{sec_research_intersection}

\begin{figure}[ht]
    \centering
    \includegraphics[trim=0 0 0 0,clip,width=\columnwidth]{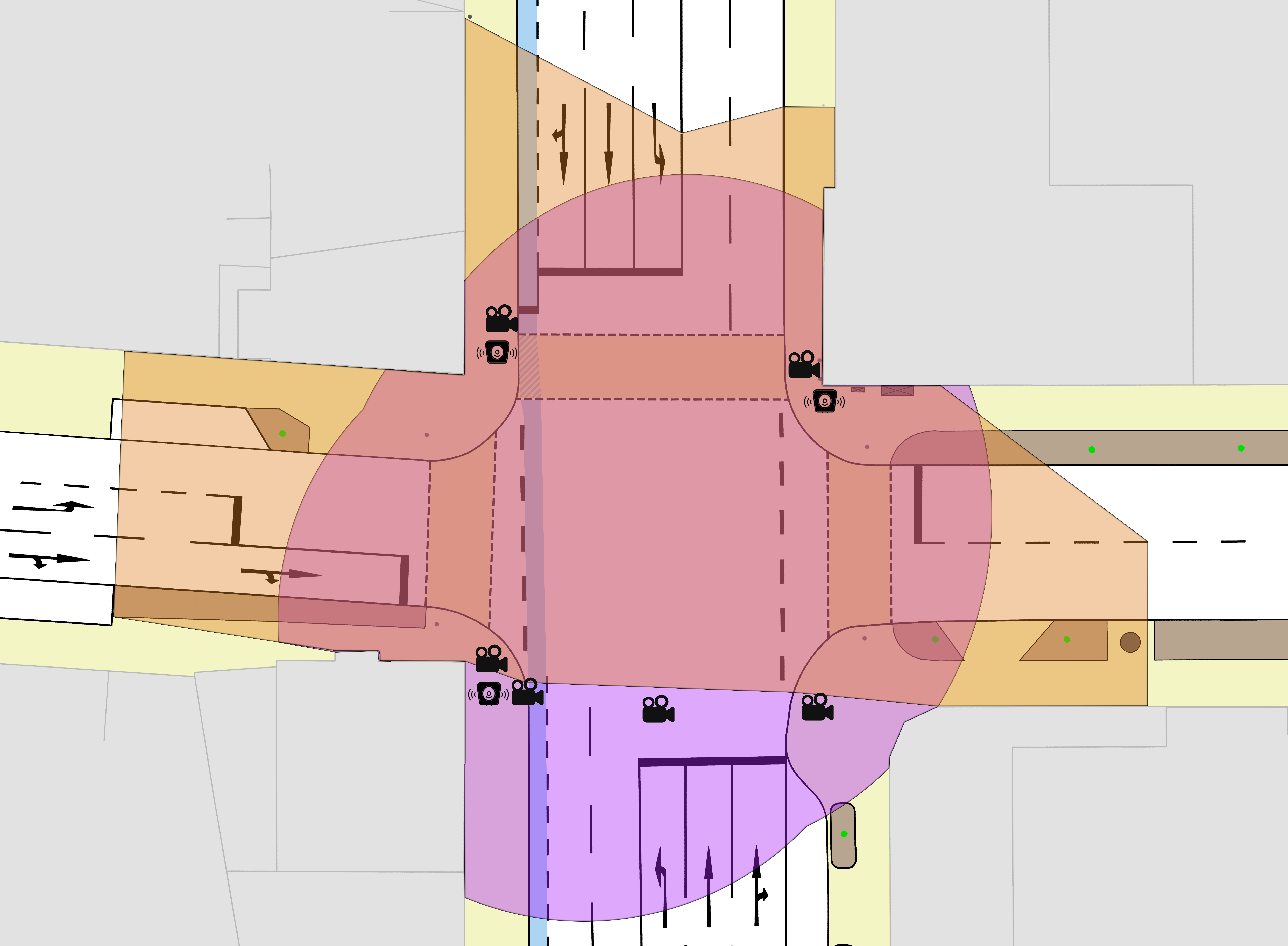}
    \caption{Illustration of the intersection topology with sensor positions (black symbols), full stereo camera coverage (orange area), and LiDAR coverage (purple area).}
    \renewcommand{\the}{}
    \label{fig:stereo_fov}
\end{figure}

A detailed description of the research intersection can be found in~\cite{xung}. The research intersection is located in the inner-city of Aschaffenburg and is highly frequented by VRUs and vehicles. The intersection is signalized and includes three pedestrian crosswalks and an additional bike lane. The speed limit is 50 km/h. The intersection has six high-resolution wide-angle color cameras (4096x2160 pixels, 71-degree aperture angle), three high-resolution spinning LiDAR sensors (128-Layers, 45-degree aperture angle), alongside a weather station, and a traffic light signal tracker for contextual data. The cameras form a correlated multi-camera stereo system and are mounted approximately eight meters above ground level to reduce occlusions. The LiDARs are mounted on light poles at five meters in height at three corners of the core intersection. The camera systems' stereo coverage, the LiDAR sensors coverage, and the topology of the intersection are illustrated in \autoref{fig:stereo_fov}. The sensor setup covers a total area of 50x40 meters for 3D road user tracking. The camera setup is used to achieve highly accurate VRU trajectories and to extract visual context information. The LiDAR setup handles all common types of road users.

%-----------------------------------------------------------------------------------------------
\subsection{Detection and Classification}
\label{sec_detection_classification}

The reliable detection of road users in camera images is essential. Therefore, we extended and fine-tuned a  Faster-RCNN R101 FPN model~\cite{faster_rcnn} with additional classes using 10,000 human annotated and equally distributed camera images, split into 80:20 for training and evaluation. To ensure variability, the images were collected at various seasons, weather conditions, and day-times. The model focuses on VRUs. Therefore, we split VRUs into sub-classes, e.g., pedestrians, cyclists, e-scooter riders, strollers, and wheelchair users. As a result, the model achieves an mAP score of 90.5\% within our static setup.

\begin{figure}[h!]
    \centering
    \includegraphics[clip,width=\columnwidth]{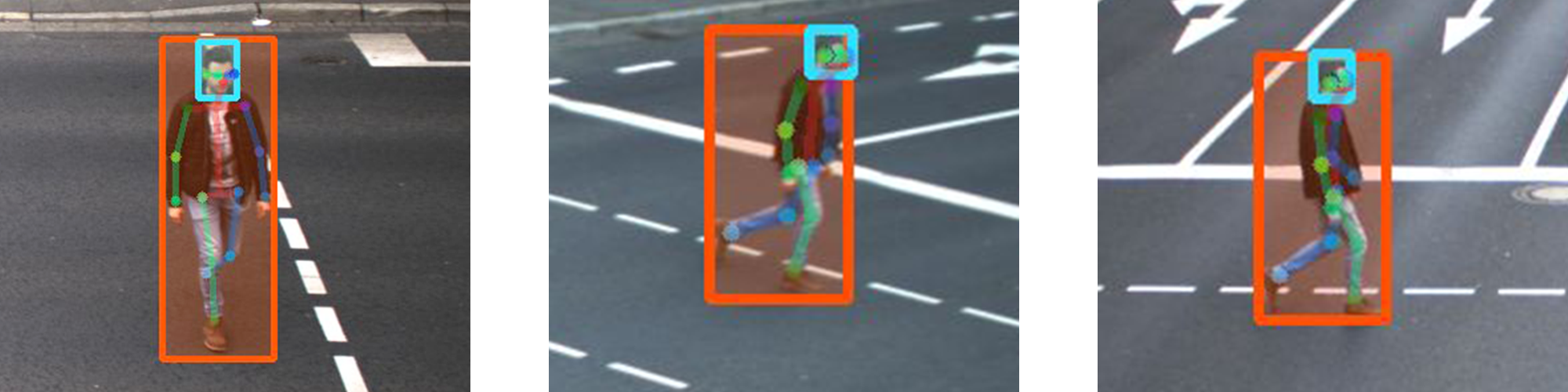}
    \caption{Exemplary VRU detection from different cameras. The red box represents the pedestrian, cyan the head detection. The green and blue dashes show the human key point and body pose detection.}
    \label{fig:multiple_vru_dets}
\end{figure}

Intrinsic and extrinsic camera calibration is necessary to achieve accurate world object coordinates from 2D detections. After mounting, a checkerboard pattern was used for the intrinsic calibration of the cameras. For the extrinsic calibration, we scattered 61 global geographic referenced points at the corners of lane markings and other road marks within the intersection area. Due to the bright white color and high reflectivity, they are easy to see in camera images and LiDAR point clouds. The global geographic referenced points are provided in Universal Transverse Mercator (UTM) coordinates, and the height is meters above sea level. For the cameras, we used the methods described in~\cite{hartley_2004}\cite{openCV3_kaehler} to solve Perspective-n-Point (PnP) pose computations to obtain the sensor's extrinsic parameters concerning an origin point. One marker is treated as a reference point to define a local coordinate system, further called intersection coordinate system. Although we were careful to perform the calibration process with maximal accuracy, we discovered inconsistencies moving from one stereo camera setup to the next. \autoref{fig:multiple_vru_dets} shows a person crossing the street being visible in multiple camera views. Labeling the center of the head of the person in every image and calculating the 3D positions concerning the stereo settings lead to a difference exceeding the expected error, especially at the margins of the image areas. This effect causes detections of the same object not to be considered corresponding between stereo camera setup changes.

Our solution is to optimize the extrinsic camera parameters to achieve a minimal re-projection error concerning the measured intersection points and minimize the pairwise distance of triangulations of corresponding points in the common fields of view. For $\mathbb{G}$ being the set of geo-referenced points, $\gamma_i(\mathbb{g})$ being a manually labeled image point in camera $i$ corresponding to $\mathbb{g}\in\mathbb{G}$, and $\pi_{i}$ being the projection of a real-world point to camera $i$ with $i\in \{1,\dots,6\}$ based on the calibration, then
\[
\epsilon_r := \sum_{\mathbb{g}\in\mathbb{G}} \sum_{i \in \{1,\dots,6\}} \Vert \pi_i(\mathbb{g}) - \gamma_i(\mathbb{g}) \Vert 
\]
is the re-projection error. Additionally, we collect a set of real-world points $\mathbb{H}$ we can see in some or all of the cameras, but we do not know their coordinates. An example is the center of the head of the person crossing the street in \autoref{fig:multiple_vru_dets}. Next, we label the corresponding positions $\gamma_i(\mathbb{h})$ of such an $\mathbb{h}\in\mathbb{H}$ with $i\in I(\mathbb{h})$ and $I(\mathbb{h}) \subset \{1,\dots,6\}$ being the cameras that have a view of $\mathbb{h}$. The triangulation $\tau_{i,j}(p,q)$ calculates the 3D world point for two points $p$ from camera $i$ and $q$ from camera $j$ assuming $(i,j)$ being a calibrated stereo camera setup. We formulate the consistency error $\epsilon_c$ AS the summed-up distance of the triangulations of the same real-world points viewed by different stereo setups.
\begin{comment}
The aforementioned set of stereo camera setups shall be called $\mathbb{S}$ and $\mathbb{S}_{I(\mathbb{h})}$ shall be the set of stereo setups containing only cameras from $I(\mathbb{h})$ of a 3D point $\mathbb{h}$. Our consistency term is the following: 
\begin{align*}
\epsilon_c =
\sum_{\mathbb{h}\in\mathbb{H}} \sum_{\substack{(i,j)\in\mathbb{S}_{I(\mathbb{h})},\\ (i',j')\in\mathbb{S}_{I(\mathbb{h})}}} &\Vert \tau_{i,j}\left(\gamma_i(\mathbb{h}),\gamma_j(\mathbb{h})\right)\\
&- \tau_{i',j'}\left(\gamma_{i'}(\mathbb{h}),\gamma_{j'}(\mathbb{h})\right) \Vert.
\end{align*}
\end{comment}
To smooth the transition between the camera stereo setups, we choose the elements of $\mathbb{H}$ to be in intersecting areas of at least two stereo setups. We use Nelder-Mead~\cite{nelder_mead} optimization on the extrinsic parameters of every camera to minimize our objective function 
\[\epsilon_r + \lambda \epsilon_c.\]
Depending on $\lambda$, the consistency argument is more or less important than the fit of the measured intersection points. In our case, $\lambda = 0.04$ is a good choice. We can achieve a precise and consistent multi-camera calibration setup by following this procedure. The mean distance of the triangulations $\tau_{i,j}(\gamma_i(\mathbb{g}),\gamma_j(\mathbb{g}))$ over all stereo camera sets $(i,j)$ and geo-referenced points $\mathbb{g}$ to the points $\mathbb{g}$ is \SI{3.4}{cm} and the maximum distance is \SI{9.6}{cm}.

\begin{figure}[h!]
    \centering
    \includegraphics[clip,width=\columnwidth]{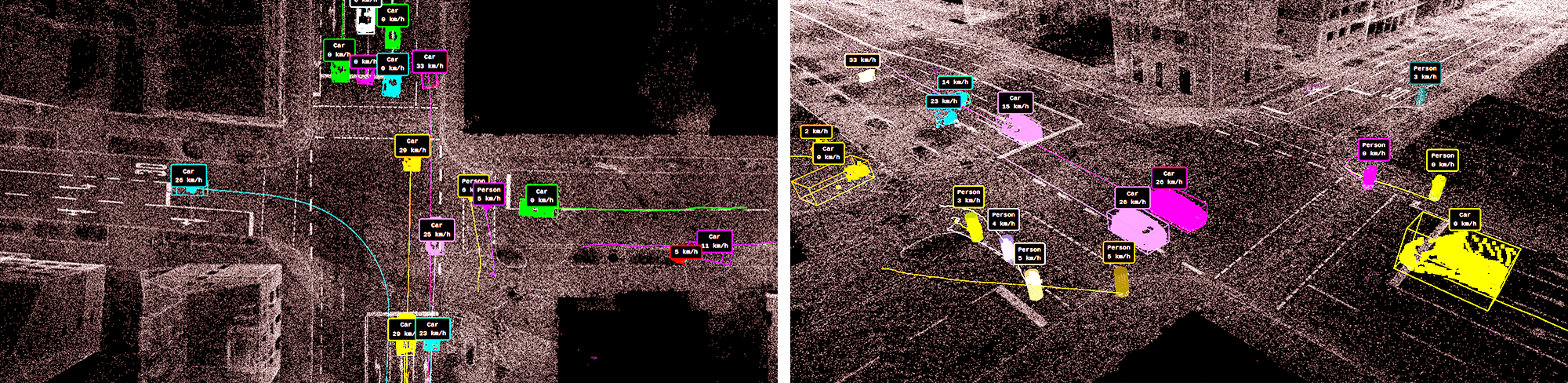}
    \caption{Exemplary LiDAR detection and tracking results. Each object is represented by a different color.}
    \label{fig:marker}
\end{figure}

The LiDAR setup consists of three sensors. We use a two-step data post-processing pipeline to achieve highly reliable 3D object trajectories. First, each sensor's data is processed for itself, using foreground-background subtraction combined with a classifier. In our case, this method achieves excellent results because of our static environment. The subtraction uses an exact digital scan of the intersection as a reference. Second, both sensor results are merged using least-squares fitting of 3D point sets~\cite{arun_1987}. As a result and in contrast to our vision system, we receive trajectory data for all road users. The reference point cloud and two exemplary results are illustrated in \autoref{fig:marker}.

%-----------------------------------------------------------------------------------------------
\subsection{Tracking and Post-processing}
\label{sec_tracking_postprocessing}

To perform further work, such as movement prediction, it is essential to have both detections and consistent and reliable tracks available. The aim of tracking is twofold. On the one hand, consistent identifiers are assigned; on the other hand, missed detections due to failures or short-time occlusions can be bridged. For this purpose, it is essential to have a fitting movement model for the tracked object. VRUs are not only pedestrians but, in any case, including persons. We present a way of tracking pedestrians, cyclists, e-scooter riders, strollers, and wheelchair users as VRU sub-classes incorporating individual movement models and class probabilities. For pedestrians, we use an Interacting Multiple Model (IMM) Kalman filter that is flexible enough to cover the capability of abruptly changing the movement direction and rapidly changing accelerations. The model state space is $[x, \dot{x}, y, \dot{y}, z, \dot{z}]$. Cyclist, e-scooter rider, and wheelchair user movements follow arcs or straight lines. They differ in acceleration and average speed. For all these classes, we use adjusted parameters. The procedure for all sub-classes is the same. We describe the method for cyclists in~\cite{reitberger}, going further in detail about how we initialize tracks and assign measurements.
\begin{comment}
Therefore, we choose the so called bicycle model, adapted from~\cite[Chapter 10]{bar_shalom_2001} based on the state space $[x, y, z, \dot{z}, \gamma, \dot{\gamma}, v]$ with $x$, $y$, and $z$ being world coordinates, $\gamma$ being the yaw, $v$ the velocity, and $\dot{\fbox{}}$ being the time derivative. The state transition is given by $f(\mathbf{x}) := [
x + cos(\gamma)\, a - sin(\gamma)\, b,
y + sin(\gamma)\, a + cos(\gamma)\, b,
z + \dot{z}\, T, \dot{z}, \gamma + \dot{\gamma}\, T, 
\dot{\gamma}, v]$
with $a =\frac{sin(\dot{\gamma}\, T)\, v}{\dot{\gamma}}$ and $b = \frac{(1 - cos(\dot{\gamma}\, T))\, v}{\dot{\gamma}}$ for a time step $T$. 
\end{comment}

The detection of a cyclist consists of an intersection of a bicycle with a person bounding box exceeding a predefined Intersection over Union (IoU) threshold score. 
\begin{comment}
Performing tracking of cyclists solely based on cyclist detections does not perform well, as especially the bicycle detection is not yet stable with regard to all viewing angles, for example.
\end{comment}
With Neural Networks (NN) based detection algorithms, we discovered an absolute bicycle detection rate difference between frontal and side views of about $0.65$. Therefore, bicycle detections need to be more robust in certain constellations that bicycle detection-only tracking cannot provide gap-free results. We tackle this issue by tracking pedestrians and bicycles simultaneously and introducing a class probability. The algorithm we selected to achieve the goal mentioned above is the Interacting Multiple Model (IMM)~\cite{imm_blair_shalom}\cite{imm_genovese} approach based on the multiple Kalman filter models.
\begin{comment}
It shows a robust behavior with respect to model mismatching~\cite{pitre_2005}. To make the different model states compatible, the IMM state is a lifted one by merging the individual state spaces~\cite{imm_diff_models_states}. 
\end{comment}
A probability score evaluates every tracking step for how well each set of models, i.e., pedestrian and bicycle models, fits the perception. Additionally, the object class predictions by the NN classifier are available. We combine both indicators to label the class of the IMM-tracked object. Together, we achieve a classification precision of at least 0.97\%. Concerning the MOTA and MOTP tracking scores~\cite{bernardin_2008}, the IMM approach can increase cyclist tracking accuracy (MOTA) by adding person detections from 32.5\% to 97.6\%. Furthermore, the precision (MOTP) improves from \SI{15.5}{cm} to \SI{7.6}{cm} by mixing in the pedestrian models.

%-----------------------------------------------------------------------------------------------
\subsection{Context Data}
\label{sec_context_data}

To achieve highly reliable VRU predictions, additional information about the current situation is essential. Therefore, we use additional sensors to perceive different types of context data in real-time. For example, a weather station provides features like temperature, wind, precipitation, and visibility. A traffic light signal tracker provides all traffic light signal statuses. Furthermore, the camera system can extract VRU human body poses, providing additional information for viewing direction detection and gesture recognition. In addition, the intersection was measured by combining photogrammetry and road-level laser scans resulting in a highly accurate digital model with a better than \SI{1}{cm} textural resolution and a \SI{3}{cm} or better structural resolution. The LiDAR reference map and an exact Open Street Map (OSM) are derived from this model. Besides VRU and other road user trajectories, our setup provides a comprehensive set of additional context information to achieve a precise environmental perception of every traffic situation.

%-----------------------------------------------------------------------------------------------
\subsection{Dataset Format and Tools}
\label{sec_format_tools}

Our goal is to ensure that the dataset is easy to use. Therefore, we provide all data synchronized and in the established Json-format. A global keyword catalog will enable fast scene browsing, including the number of tracks, weather-, lightning-, and seasonality attributes. Every scene will include the following scope:
\newline
\textbf{High-level data:} A timestamp-ordered Json-file including all detected and classified objects represented by their current 3D world coordinates and corresponding meta attributes like detection scores. Furthermore, all weather- and traffic light signal data is included. Next, the intersection topology is given by a precise OSM reference map. The map represents all static circumstances. Finally, additional visual support is given by a scene preview video for a better scene understanding. All six camera images and a trajectory top-view presentation are stitched together for the preview. \autoref{fig:scene_preview} depicts a sample frame of a scene preview video.
\newline
\textbf{Low-level data:} Object detection lists and human key point annotations are provided for every frame. Furthermore, the LiDAR sensors' point clouds will be included. Every point is described by 3D world coordinates and its reflectivity $(x, y, z, r)$. In addition, object detection lists are provided for every point cloud. Finally, the extrinsic and intrinsic camera parameters are included, just as all 61 survey points. The low-level data can be used to extract additional context information from sensor data if necessary and not already provided.

\section{\large Dataset}
\label{sec_dataset}

\begin{table*}
\caption{Comparison of existing VRU focused trajectory datasets.}
\label{tab_datasets_comparison}
\begin{tabularx}{\textwidth}{@{} l *{10}{C} c @{}}
\toprule
\textbf{Set} & \textbf{Type} & \textbf{VRUs} & \textbf{Vehicles} & \textbf{Classes} & \textbf{Method} & \textbf{FPS} & \textbf{Track Length} & \textbf{Context} & \textbf{Weather}  & \textbf{Seasonality}\\ 
\midrule
\textbf{SSD} & drone & 9.7K & 0.6K & 6 & camera & 25 Hz & 20 s & no & clear & no \\
\textbf{CITR+DUT} & drone & 2.1K & 0.1K & 2 & camera & 24 Hz & 25 s & no & clear & no \\
\textbf{inD} & drone & 5K & 6.5K & 5 & camera & 25 Hz & 45 s & no & clear & no \\
\textbf{DeCoint} & infrastructure & 3.3K & 0 & 2 & camera & 25 Hz & 30 s & no & clear & no \\ 
\textbf{PIE} & vehicle & 1.8K & 0 & 5 & camera & 30 Hz & 13 s & yes & clear & no \\
\textbf{JAAD} & vehicle & 2.8K & 0 & 1 & camera & 30 Hz & 4 s & yes & mixed & yes \\
\textbf{PVI} & vehicle & 6.1K & 0 & 3 & camera+lidar & 10 Hz & 10 s & yes & mixed & no \\
\midrule
\textbf{IMPTC} & infrastructure & 2.5K & 20K & 8 & camera+lidar & 25 Hz & 60 s & yes & mixed & yes \\
\bottomrule
\end{tabularx}
\end{table*}

After describing our research intersection and data processing pipeline in the previous section, the following chapter shifts the focus toward the IMPTC dataset. First, \autoref{sec_glance} will give a detailed dataset overview. Afterward, \autoref{sec_comparison} compares our dataset with existing ones.

%-----------------------------------------------------------------------------------------------
\subsection{IMPTC at a Glance}
\label{sec_glance}

\begin{figure}[h]
    \centering
    \includegraphics[trim=0 90 0 90,clip,width=\columnwidth]{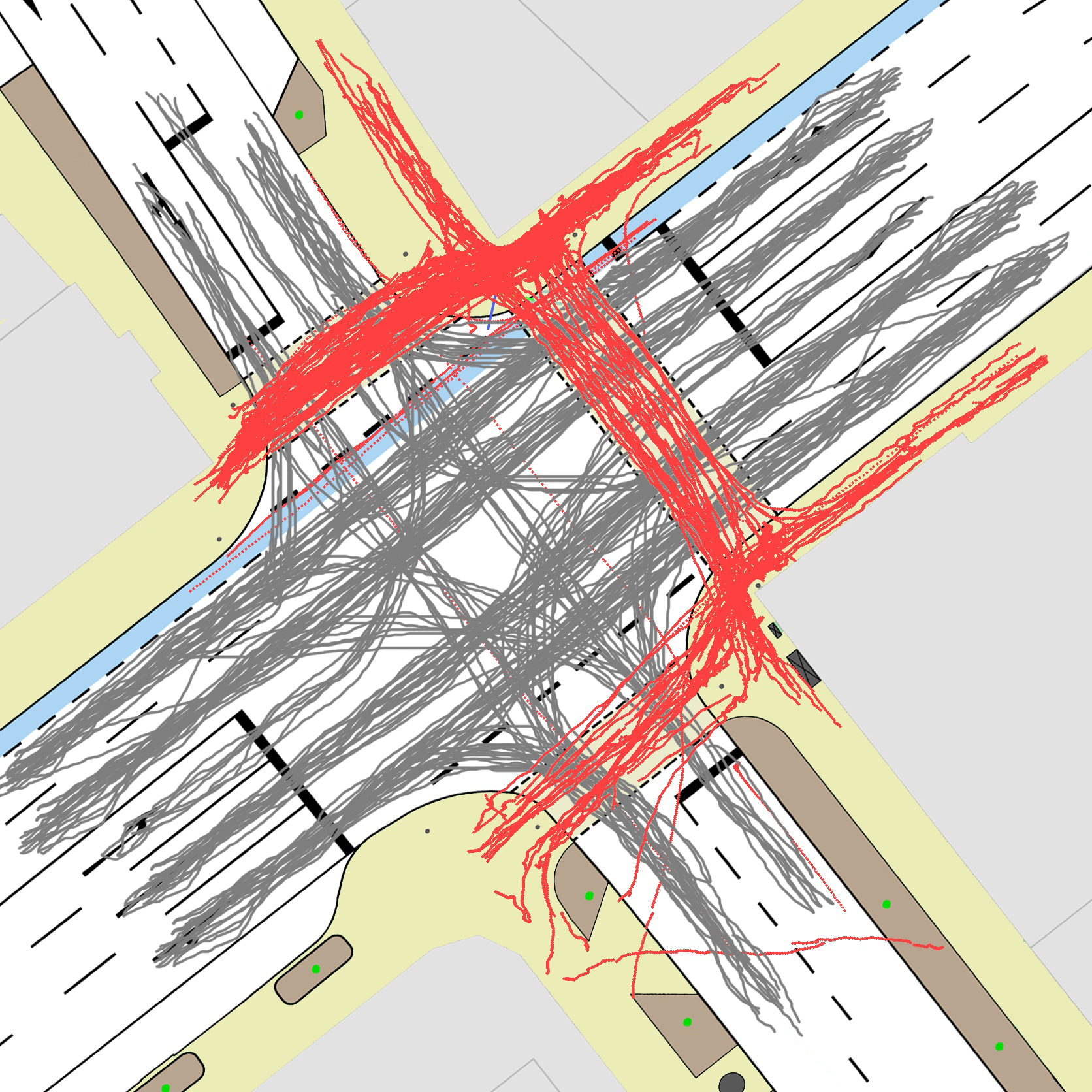}
    \caption{One hundred random extracted VRU trajectories (red) and corresponding vehicle tracks (gray) from the IMPTC dataset.}
    \label{fig:multiple_tracks_on_map}
\end{figure}

In \autoref{sec_state_of_the_art}, we elaborate that there is a clear need for more detailed and complete datasets in VRU behavior prediction and trajectory forecasting, especially in critical urban traffic situations like intersections. It is the goal of our dataset to fill this gap. Therefore, we provide one of the most extensive trajectory datasets so far. IMPTC will be released with an initial number of 250 sequences recorded in 2022 and beyond. Each sequence has an average length of 90 to 120 seconds and represents public everyday traffic situations at our intersection. The dataset includes 2,500 VRU- and 20,000 vehicle trajectories. VRUs are classified into subgroups: pedestrians, cyclists, e-scooter riders, strollers, and wheelchair users. \autoref{fig:multiple_tracks_on_map} illustrates 100 randomly selected VRU trajectories and corresponding vehicle tracks. Vehicles are divided into cars, motorbikes, and trucks/buses. An unknown class is also included representing uncommon corner cases. Besides object classification and 3D trajectories, IMPTC contains multiple types of additional context information and low-level sensor data like LiDAR point clouds as described in \autoref{sec_context_data} and \autoref{sec_format_tools}. The initial set will be extended over time to enlarge the dataset's variability, including different daytime and light conditions, weather and seasonality conditions, and VRU amounts and types. The dataset's goal is to support researchers within the topics of VRU trajectory and behavior/intention prediction, traffic flow understanding, and social-behavioral analysis in complex urban traffic scenarios.

%-----------------------------------------------------------------------------------------------
\subsection{Comparison with Existing Datasets}
\label{sec_comparison}

In \autoref{tab_datasets_comparison}, we compare the IMPTC dataset with currently available VRU-focused datasets: SSD~\cite{ssd}, CITR+DUT~\cite{citr}, DeCoint~\cite{decoint} inD~\cite{inD}, PIE~\cite{pie}, JAAD~\cite{jaad}, and PVI~\cite{euro_pvi}. No dataset meets all requirements. A comprehensive roadside VRU trajectory dataset should include a balanced number of road user trajectories at different weather, lightning, and seasonality conditions. Furthermore, additional context information should be included to create the best environmental perception model possible as the basis for VRU behavior prediction. Regarding total available VRU trajectories, our initial set can not match inD, PVI, or SSD. Nevertheless, IMPTC covers the broadest range of object classes, eight in total, with five VRU subclasses, including new means of transportation like e-scooters. Only a few compared datasets include a wide range of weather or seasonality. Sunny and cloudy weather is omnipresent. Thanks to our LiDAR sensors, we can record data under poor weather and lighting conditions and at night. PIE, JAAD, and PVI provide additional context information but with varying scopes. In \autoref{sec_context_data}, we detailed the context information included in IMPTC, exceeding all others. Our dataset is the most balanced and extensive roadside VRU dataset yet. All information, download links, detailed descriptions, instructions, and additional code will be available at \textbf{\url{https://github.com/kav-institute/imptc-dataset}}.
\section{\large Conclusion}
\label{sec_conclusion}

This paper has motivated the need for an extensive VRU trajectory dataset at urban intersections, which other publicly available datasets still need to meet. We have shown that the available datasets use different recording strategies, e.g., static setups, drones, or research vehicles, and vary in size and scope of provided trajectory data. So far, these datasets have yet to take care of additional important context information, i.e., static circumstances, weather, traffic light signals, or VRU body poses, necessary for reliable VRU behavior prediction. Furthermore, we subdivide VRUs into five groups, i.e., pedestrian, cyclist, e-scooter rider, strollers, and wheelchair user, for a more precise analysis of the movement behavior. Our dataset fills that gap and provides researchers with the densest environmental perception available. After implementing a complete processing pipeline, we used that pipeline to create the IMPTC dataset. The initial set contains 250 scenes with over eight hours of recorded data and 2,500 VRU trajectories. The set will be continuously extended. We surpassed any comparable dataset regarding data scope in high-level and low-level aspects and outlined multiple application domains in which the IMPTC dataset supports researchers. The dataset will be released after the conference date, and it will be updated at regular intervals.

\end{document}